# Consistency-Driven Dual LSTM Models for Kinematic Control of a Wearable Soft Robotic Arm*

Xingyu Chen, Yi Xiong, and Li Wen, *Senior Member, IEEE*

*Abstract*— In this paper, we introduce a consistency-driven dual LSTM framework for accurately learning both the forward and inverse kinematics of a pneumatically actuated soft robotic arm integrated into a wearable device. This approach effectively captures the nonlinear and hysteretic behaviors of soft pneumatic actuators while addressing the one-to-many mapping challenge between actuation inputs and end-effector positions. By incorporating a cycle consistency loss, we enhance physical realism and improve the stability of inverse predictions. Extensive experiments—including trajectory tracking, ablation studies, and wearable demonstrations—confirm the effectiveness of our method. Results indicate that the inclusion of the consistency loss significantly boosts prediction accuracy and promotes physical consistency over conventional approaches. Moreover, the wearable soft robotic arm demonstrates strong human–robot collaboration capabilities and adaptability in everyday tasks such as object handover, obstacle-aware pick-and-place, and drawer operation. This work underscores the promising potential of learning-based kinematic models for human-centric, wearable robotic systems.

## I. Introduction

Soft robotic arms are typically composed of pneumatic chambers, rod-driven, or tendon-driven mechanisms, and are designed for applications that demand safety in human–robot interaction as well as operation in constrained environments. Several inherent characteristics of soft robotic systems make them particularly well-suited for wearable applications on the human body[1].

In rigid robots, the kinematic characteristics of conventional six-degree-of-freedom rigid manipulators are well-defined and relatively straightforward to model[2], [3], [4]. Their actuators also exhibit highly characteristic and predictable output properties. To ensure safety in human–robot interaction, this study employs pneumatic actuation. In contrast, for soft robotic arms, the degrees of freedom of the continuum segment are more than those of the input actuation, resulting in a non-bijective mapping between actuation inputs and end-effector positions—that is, a distinct one-to-many relationship. Furthermore, for a single pneumatic soft actuator, the viscoelasticity and delayed response of the soft material cause the pressure–length relationship to exhibit pronounced hysteresis, manifesting as a typical closed-loop curve[5]. This relationship is strongly dependent on actuation history, which introduces significant challenges to both modeling and control.

The constant curvature (CC) model is the most widely used control strategy for soft robots, assuming uniform curvature along the manipulator to achieve low computational cost in steady-state control [6]. Extending this, the piecewise constant curvature (PCC) model represents the robot as multiple CC segments [7], but it only maps actuator lengths to poses, failing to capture actuator hysteresis and making inverse kinematics difficult. Consequently, machine learning methods have been introduced for soft arm modeling and control. Neural networks have been applied to both forward and inverse kinematics [8], with LSTM and RNN models capturing temporal dependencies [9], [10], and other methods using loss design or numerical optimization for inverse solutions [11]. However, most train forward and inverse models separately, lacking consistency constraints that limit inverse accuracy and stability.

To address the aforementioned challenges, this study proposes a neural network modeling framework that enforces forward–inverse consistency by leveraging temporal information. Specifically, we first construct an LSTM-based forward model to predict the mapping from pneumatic pressure inputs to end-effector poses, while simultaneously designing an inverse model to estimate the actuation inputs required to achieve a desired target position. To enhance the consistency between the two models and improve the stability of the inverse model, we introduce a forward–inverse consistency loss during training[12], which enforces alignment between the forward predictions and inverse outputs in the latent representation space.

Meanwhile, we propose employing the soft robotic arm as a wearable assistive device to aid users with impaired hand function in performing grasping tasks and operations within confined environments. Existing wearable soft devices have primarily focused on rehabilitation[13], [14], assistance[15], [16], prosthetics[17], [18], and therapeutic purposes[19], [20], serving mainly to assist users with neuromuscular disorders in performing daily tasks. By contrast, applications involving supernumerary wearable devices are relatively limited[21], [22], [23], [24]. As a "third hand" for humans, soft robotic arms can execute highly dexterous operations. Therefore, we propose the use of a soft robotic arm as a wearable human-assistive device to support individuals with impaired hand functionality in performing grasping tasks as well as operations in confined environments.

This work presents a temporal-information-based model architecture and a training paradigm incorporating a forward–inverse consistency loss. Validation experiments, including trajectory tracking and accuracy comparison, demonstrate the superiority of the proposed approach. A wearable soft robotic arm system was also developed to verify its effectiveness and application potential. Section 2 introduces the hardware design, control challenges, and model

*This work was supported by the National Key R&D Program of China (2024YFB4707300) and the National Science Foundation support projects, China (Grant Nos. 62425303, T2121003 and 62403029).

The authors are with the School of Mechanical Engineering and Automation, Beihang University, Beijing 100191, China (e-mail: xingyuchen@buaa.edu.cn; 21376267@buaa.edu.cn; liwen@buaa.edu.cn).

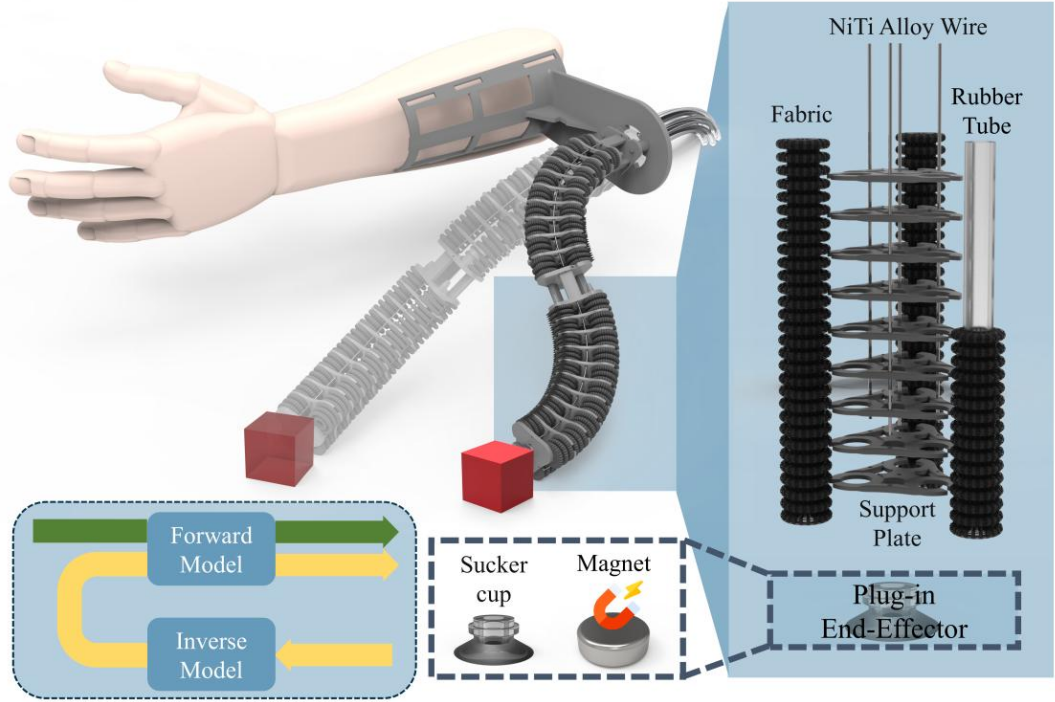

Figure 1. Overview of wearable soft robotic arm with actuator structure and model framework

training methodology. Section 3 presents the trajectory tracking and accuracy comparison results, along with wearable demonstrations. Section 4 discusses potential applications and future work.

## II. METHODS

### A. Soft Robotic Arm Design

The soft robotic arm is pneumatically actuated using McKibben pneumatic artificial muscles, each composed of an elastic silicone tube encased in a corrugated fabric sleeve that limits radial expansion to allow only axial elongation. The actuators exhibit high extensibility (up to 202%) and low weight, providing sufficient rigidity when unpressurized for stable horizontal support.

The soft robotic arm comprises two serial segments, each formed by three parallel actuators, providing six degrees of freedom. Spacer plates evenly separate and fix the actuators to enhance structural support. A central shape memory alloy (SMA) wire constrains the arm's length, allowing bending without axial extension. To counteract instability from actuator extensibility, three additional SMA wires are uniformly arranged around each segment to limit torsion and improve stability.

The robotic arm is attached to the user's forearm via a 3D-printed bracket and elastic straps. With the forearm horizontal, the soft arm assists in grasping and placing objects using a suction-cup or magnet end effector. Its six pneumatic channels are controlled by proportional valves.

### B. Dataset Collection Method

To train and validate the proposed kinematic model, a randomized sequential data collection strategy was developed. A three-dimensional actuation vector satisfying a fixed-sum constraint was randomly generated and expanded to a six-dimensional control input. Linear interpolation between target frames ensured smooth control signals, with zero-mean noise added to enhance data diversity. The sampling frequency was set to one frame every 3 seconds to maintain quasi-static actuation of the soft robotic arm, resulting in a total of 13,000 frames of data. The actuation pressure ranged from 0 to 250 kPa. The control signals were then applied to the pneumatic actuators while recording timestamps and end-effector poses using a motion capture system (NOKOV, Beijing, China), yielding a temporally rich training dataset.

### C. Problem Analysis

*Nonlinear & Hysteretic Actuator Behavior:*

For the McKibben actuators used in this study, the pressure–length relationship exhibits a pronounced hysteresis loop: during loading, a higher pressure is required to reach a given length, whereas during unloading, the actuator returns to the same length under a comparatively lower pressure, as shown in Figure 2. This phenomenon primarily arises from the combined effects of material properties, structural factors, and pneumatic dynamics. First, the elastomeric bladder possesses inherent viscoelasticity, leading to energy dissipation during loading–unloading cycles. Second, friction between the bladder and the braided sleeve, together with fiber rearrangement and changes in braid angle induced by pressure variations, introduces geometric path differences. In addition, flow resistance in the pneumatic system, valve response delays, and asymmetry between inflation and deflation processes further exacerbate this hysteresis effect. The interplay of these mechanisms results in non-overlapping loading and unloading curves, thereby producing a characteristic hysteresis loop in the actuator's pressure–length profile, as illustrated in Figure 2(b).

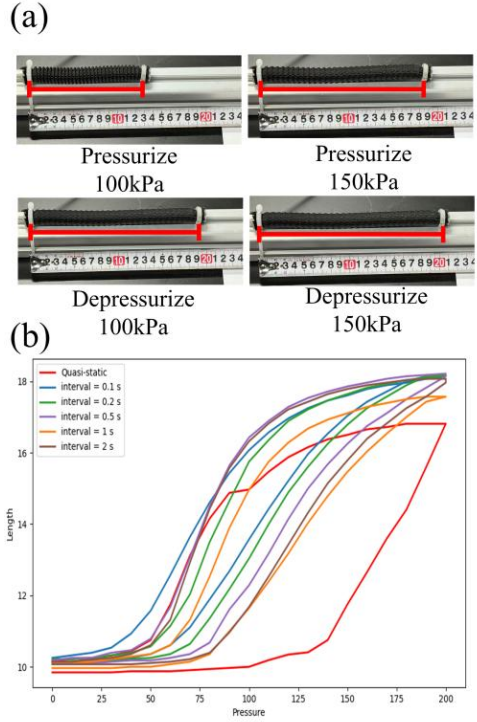

Figure 2. Illustration of the hysteresis characteristics of the soft pneumatic actuator. (a) Length difference of the actuator between pressurization and depressurization under the same pressure. (b) Length–pressure curves of the actuator under quasi-static conditions and different time intervals with equal pressure step size.

Due to this phenomenon of the actuators, when we need to predict the length and pose of the soft actuators and the soft robotic arm, or inversely predict the pressure of the soft robotic arm, it is necessary to consider the temporal information of the pose or pressure in order to determine which state the actuators are currently in within the hysteresis loop.

In addition, as shown in Figure 2(b), under equal step-size input of increasing and decreasing pressure, different time intervals of the steps also result in different shapes of the hysteresis loop. This indicates that the behavior of the actuators is highly dependent on their dynamic characteristics. Therefore, when controlling the soft robotic arm, the step size and time interval should be kept as consistent as possible, which at the same time emphasizes the importance of using time-series dependent models.

### D. Model Architecture

In this study, two neural network models were constructed, corresponding to the forward kinematics modeling and inverse kinematics modeling of the soft robotic arm as illustrated in Figure 3.

The forward model employs a two-layer LSTM network followed by an FC–ReLU layer and an FC–Sigmoid layer to generate the final output. The input comprises pressure values and their variations from the past two time steps to the target step, while the output is the predicted end-effector pose at the next step. Incorporating pressure variations highlights the system's differential dynamics, improving nonlinear mapping accuracy, and the Sigmoid activation ensures physically valid predictions.

In the inverse model, a two-layer LSTM followed by a fully connected ReLU layer is employed. The model input consists of three components: (1) the repeated target position increments, aligned with the temporal sequence; (2) the historical sequence of target position increments; and (3) the current sequence of target position increments. The model outputs the driving pressures required to achieve the given target pose. To emphasize the guiding effect of the target pose on pressure prediction, a task prior is incorporated into the initialization of the LSTM's hidden states. Specifically, the target pose is mapped into the network's latent space through $(h_0, c_0) = \tanh(W \cdot \text{target}_{xyz} + b)$, thereby enhancing the influence of target information during temporal sequence modeling and improving the convergence performance of the model in sequence learning.

Both the forward and inverse models use two-layer LSTM networks with a hidden size of 32 and a fully connected (FC) hidden size of 16. The overall training batch size is set to 32.

In addition, to eliminate the influence of dimensional differences among different physical quantities on training convergence, all input and output data in this study were normalized using z-score standardization. This normalization method ensures zero mean and unit variance of the data distribution, thereby improving the stability and generalization capability of the network training process.

### E. Cycle Consistency Loss

To ensure both the accurate fitting of system dynamics by the forward model and the physical feasibility of the control inputs generated by the inverse model, this study introduces a Cycle Consistency Loss, defined as:

$$L = L_{\text{forward}} + \lambda L_{\text{consistency}} \quad (1)$$

where $\lambda$ is a weighting hyperparameter used to balance the relative importance of the two loss components.

Let $F(\cdot)$ denote the forward model, whose input consists of the normalized historical pressure sequence and differential terms (with a sequence length of 3 in this work, including one future control step), and whose output is the normalized pose prediction $\hat{p}$. For a single sample, the forward loss is defined using the Mean Squared Error (MSE):

$$L_{\text{forward}} = \| \hat{p} - p_{\text{gt}} \|_2^2 \quad (2)$$

where $\hat{p} = F(\text{history}, \text{controls})$.

All inputs and outputs are computed within the z-score normalized space, ensuring dimensional consistency and numerical stability.

Let $G(\cdot)$ denote the inverse model, which outputs the control input given a target pose and pressure sequence. For a target pose $p^*$ (also z-score normalized), the inverse model produces a candidate control input $\hat{u} = G(p^*, \text{history})$. This candidate control is then fed into the forward model for one (or multiple) rollouts to obtain the resulting predicted pose $p_{\text{roll}}$. The consistency loss quantifies the discrepancy between this rollout result and the true target pose:

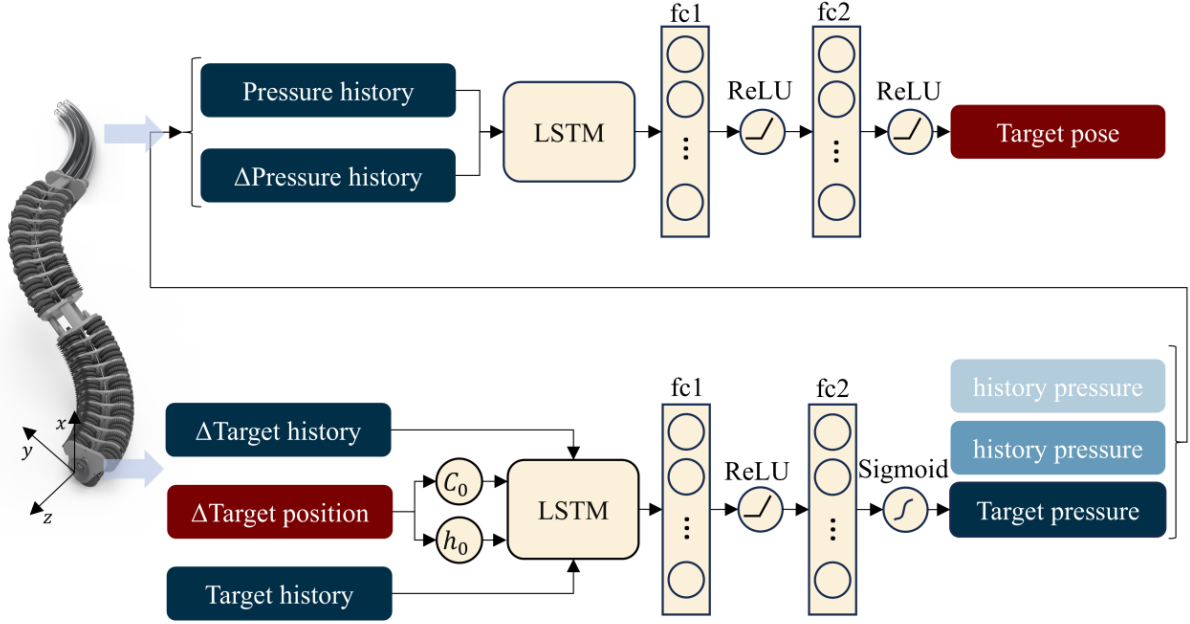

Figure 3. Machine learning model architecture for controlling the kinematic model of the soft robotic arm

$$L_{consistency}=\|p_{roll}-p^*\|_2^2, \quad (3)$$

$$p_{roll}=F(\text{history modified by } \hat{u}) \quad (4)$$

Here, "history modified by $\hat{u}$" indicates that the control signal output by the inverse model is embedded into the input window of the forward model in the appropriate format (e.g., as the pressure or differential term at the future time step), after which the forward network predicts the corresponding pose. This rollout process can be extended to a multi-step rollout to further constrain the feasibility of the inverse model's output over a longer temporal horizon.

A common difficulty in the inverse problem is the "one-to-many" relationship: for the same target position, multiple distinct control sequences may lead the system to that position. If the inverse model is trained solely through supervised regression on the control inputs (e.g., minimizing $\|\hat{u}-u_{gt}\|^2$), the model tends to learn the conditional mean $E[u|p^*]$—the so-called averaged solution. However, this mean control typically does not correspond to any physically valid control sequence and therefore fails to produce the target pose under the true forward dynamics (i.e., it is physically infeasible).

The Cycle Consistency Loss incorporates the physical constraint—represented by the forward model F—into the inverse model's training pipeline, directly evaluating whether the inverse output, when propagated through the forward dynamics, can reach the target pose. In other words, the consistency term minimizes:

$$E\left[\|F(G(p^*, \text{history}), \text{history})-p^*\|^2\right], \quad (5)$$

This term directly penalizes physically (or approximately physically) invalid averaged solutions, thereby encouraging the inverse model to select control inputs that can actually produce the target pose under F—that is, executable solutions. In this way, the approach enhances the physical consistency of the inverse model's output while alleviating the ambiguity inherent in purely supervised control regression.

III. RESULTS

To demonstrate the effectiveness of the proposed algorithm and the design of the wearable soft robotic arm, trajectory tracking experiments, accuracy comparison experiments, and wearable device demonstration experiments were conducted in this study.

A. Trajectory Tracking

To validate the effectiveness of the proposed control algorithm for the soft robotic arm, a trajectory tracking experiment was conducted. A predefined planar trajectory was mapped into the workspace of the soft robotic arm to form the mapped trajectory. The inverse kinematics model was then used to drive the arm to track this trajectory. As shown in Figure 4, the soft robotic arm successfully followed the desired trajectory. In the trajectory tracking experiments across four trajectories, the average tracking error was 31.43 mm, with a standard deviation of 13.85 mm and a maximum error of 65.86 mm.

B. Ablation Study

To evaluate the effectiveness of the consistency loss and the adverse impact of the inverse loss on model training, we compared the accuracy of models trained with different loss combinations, as shown in Table I. The results show that the model trained with the combination of forward + inverse + consistency losses performs better than the one without the consistency term, indicating the effectiveness of the consistency loss. Moreover, the model trained with the forward + consistency loss combination outperforms the one that includes the inverse loss, demonstrating that the inverse loss negatively affects model training.

TABLE I. ACCURACY UNDER DIFFERENT LOSS SETTINGS

| Loss Formulation | f + i | f + i + c | f + c |
|---|---|---|---|
| Mean Error (mm) | 43.94 | 36.03 | 30.47 |
| Max Error (mm) | 86.50 | 77.47 | 62.25 |
| Error Std. (mm) | 23.52 | 17.74 | 14.42 |

"f" denotes the forward loss; "i" denotes the inverse loss; "f" denotes the consistency loss.

## C. Wearable Pick-and-Place Demonstration

Finally, we integrated the designed soft robotic arm into a wearable device, which was mounted on the left forearm of a human subject. To validate the feasibility, adaptability, and collaborative capability of the system, a series of demonstration experiments were conducted, as illustrated in Figure 5. The experimental tasks included handing an object to a human hand, performing pick-and-place operations around obstacles, retrieving and placing objects from elevated positions, pulling open and hooking open a drawer, retrieving an object from a narrow gap, moving an object through vertical bars, passing an object through slits, and extracting an object from a pipe. Two types of end effectors—a suction-based and a magnetic type—were employed to accommodate different task scenarios. The results demonstrate that the wearable soft robotic arm can effectively operate in confined and cluttered environments, assisting humans in various daily manipulation tasks with high compliance, adaptability, and operational flexibility.

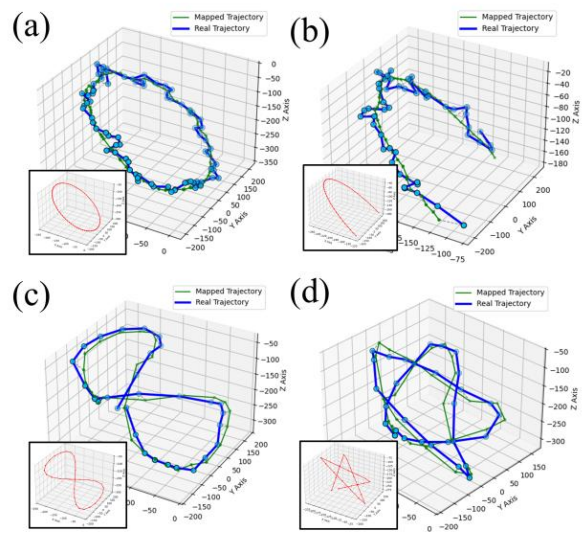

Figure 4. Trajectory tracking results of the soft robotic arm: (a) circle trajectory; (b) u-shape trajectory; (c) figure-8 trajectory; (d) star trajectory. The red trajectory represents the original planar path, the green trajectory denotes the mapped trajectory in the workspace, and the blue trajectory indicates the actual trajectory.

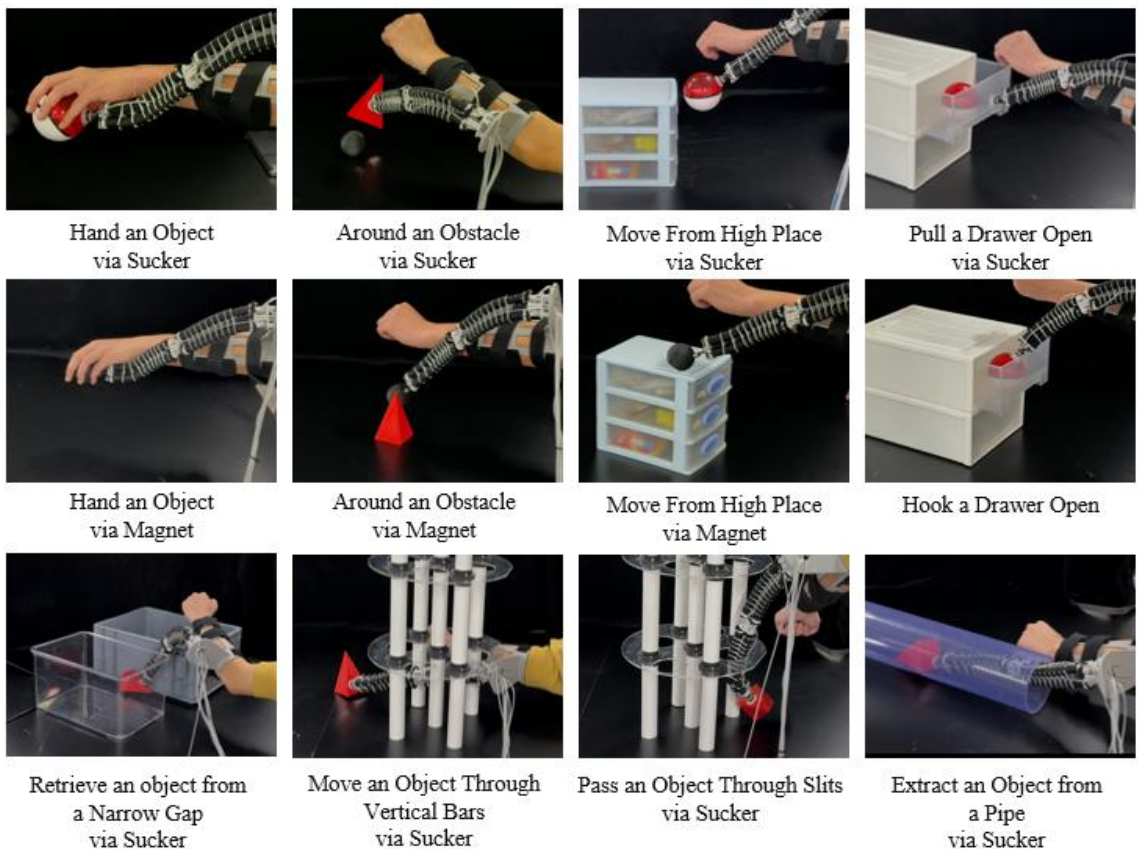

Figure 5. Schematic illustration of the wearable soft robotic arm experiment on the human body

Moreover, the "object-handing" experiment further verifies the feasibility and potential of the device for human–robot collaborative applications.

IV. CONCLUSION

This work presents a consistency-driven dual LSTM framework for learning the forward and inverse kinematics of a pneumatically actuated soft robotic arm, which is further integrated into a wearable device. The proposed framework effectively addresses the hysteresis behavior of pneumatic actuators and the one-to-many mapping problem between actuation and end-effector position, while demonstrating the potential of wearable soft robotic arms in daily task assistance. With the proposed method, the mean prediction error of the soft robotic arm was reduced from 43.94mm to 30.47mm, and a series of concise yet effective optimization strategies were implemented to further enhance the control performance of the soft arm.

For soft actuators, hysteresis and multi-solution problems are ubiquitous. Therefore, the proposed framework can be broadly applied to pneumatically actuated soft robotic arms of various configurations to optimize their control accuracy. Moreover, the model's compact structure enables onboard computation on wearable devices with limited computational resources. In the future, we plan to further improve the control accuracy and realize autonomous control of the wearable soft robotic arm, enabling more seamless human–robot collaboration.

ACKNOWLEDGMENT

The authors would like to thank Yang Liu for early discussions, Bo Yuan for her valuable suggestions, and Youning Duo for his assistance during the experiments. We also appreciate Weixi Chen and Zeyu Lou for their contributions to the initial idea validation.